\documentclass[journal]{IEEEtran}

    \usepackage{amsmath}                 % AMS Math environments
    \usepackage{amssymb}                 % Extra math symbols
    \usepackage{mathtools}               % Extra math commands and symbols
    \usepackage{mathrsfs}                % Additional math fonts

    \usepackage[square,numbers,sort&compress]{natbib} % Compressing citations
    \usepackage[capitalise]{cleveref}                 % Best reference package
    \usepackage{hyperref}                % Hypref Package for in-pdf links

    \usepackage[table,xcdraw]{xcolor}
    \usepackage{booktabs}               % Nice looking table 
    \usepackage{etoolbox}
    \makeatletter \patchcmd{\@makecaption} {\scshape} {} {} {} \makeatother
    \usepackage{color}                  % Adding shading and color

    \usepackage{algorithmicx}
    \usepackage{algcompatible}

    \algrenewcommand\algorithmicrequire{\textbf{Input:}}
    \algrenewcommand\algorithmicensure {\textbf{Output:}}

    \RequirePackage{algorithm}

    \newcommand{\algorithmicfontsize}[1]{%
     \makeatletter
     \renewcommand{\ALG@beginalgorithmic}{#1}%
     \makeatother
    }

    \newcommand{\norm}[1]{\left\Vert{#1}\right\Vert}

    \newcommand{\D}{\mathcal{D}}

    \renewcommand{\O}{\mathcal{O}}
    \newcommand{\R}{\mathbb{R}}

    \DeclareMathOperator*{\argmin}{arg\,min}

\title{Sparsity-Based Super Resolution for SEM Images}
\author{

    Shahar Tsiper$^1$, Or Dicker$^1$, Idan Kaizerman$^3$, Zeev Zohar$^3$, Mordechai Segev$^{2}$ and Yonina C. Eldar$^1$

    \thanks{
        This research was carried out in the context of the MAGNET program ``METRO 450''.
        M.S. gratefully acknowledges the support of a PoC (Proof of Concept) grant from the European Research Council.
        In addition, this work was supported by the Ministry of Science, by the ISF I-CORE joint research center of the Technion, Israel, and by the European Union’s Horizon 2020 research and innovation program under grant agreement no. 646804-ERC-COG-BNYQ.\@

        1. S. Tsiper (tsiper@technion.ac.il) O. Dicker and Y. C. Eldar (yonina@ee.technion.ac.il) are with the Department of Electrical Engineering,
        Technion --- Israel Institute of Technology, Haifa, Israel.

        2. M. Segev (msegev@technion.ac.il) is with the Physics Department and Solid-State Institute, Technion --- Israel Institute of Technology, Haifa, Israel.

        3. I. Kaizerman and Z. Zohar are with Applied Materials, 9 Oppenheimer St., Rehovot, Israel.

        The final publication is available at ACS Nano Letters via \href{http://dx.doi.org/10.1021/acs.nanolett.7b02091}{http://dx.doi.org/10.1021/acs.nanolett.7b02091}
    }
}

\begin{document}

\maketitle 

\begin{abstract}
    The scanning electron microscope (SEM) is an electron microscope that produces an image of a sample by scanning it with a focused beam of electrons.
    The electrons interact with the atoms in the sample, which emit secondary electrons that contain information about the surface topography and composition.
    The sample is scanned by the electron beam point by point, until an image of the surface is formed.
    Since its invention in 1942, the capabilities of SEMs have become paramount in the discovery and understanding of the nanometer world, and today it is extensively used for both research and in industry.
    In principle, SEMs can achieve resolution better than one nanometer.
    However, for many applications, working at sub-nanometer resolution implies an exceedingly large number of scanning points.
    For exactly this reason, the SEM diagnostics of microelectronic chips is performed either at high resolution (HR) over a small area or at low resolution (LR) while capturing a larger portion of the chip.
    Here, we employ sparse coding and dictionary learning to algorithmically enhance low-resolution SEM images of microelectronic chips up to the level of the HR images acquired by slow SEM scans, while considerably reducing the noise.
    Our methodology consists of two steps: an offline stage of learning a joint dictionary from a sequence of LR and HR images of the same region in the chip, followed by a fast-online super-resolution step where the resolution of a new LR image is enhanced.
    We provide several examples with typical chips used in the microelectronics industry, as well as a statistical study on arbitrary images with characteristic structural features.
    Conceptually, our method works well when the images have similar characteristics, as microelectronics chips do. This work demonstrates that employing sparsity concepts can greatly improve the performance of SEM, thereby considerably increasing the scanning throughput without compromising on analysis quality and resolution.
\end{abstract}


The scanning electron microscope (SEM) is one of the most versatile examination and analysis tools for solid objects at high (sub-nanometer) resolutions~\cite{Goldstein2003}.
The SEM works by launching electrons at a specimen by a focused electron beam, and then examining the emission of secondary electrons from the sample.
The secondary electrons contain topographic and compositional information about the surface.
Their flux is recorded by scanning the sample point by point.
A full image is then formed by stitching together the scanning results.
Since the first prototypes of SEM microscopes~\cite{VonArdenne1938,Zworykin1942}, and the first commercial SEM scanner~\cite{Pease1965}, many improvements have been introduced to modern SEMs.
Scanners today have faster acquisition and larger magnification, and they can store images digitally, opening a new path for real-time digital processing.
In addition, modern SEMs allow to image the surface of specimens from several perspectives simultaneously (where each perspective translates into a different viewing angle of the sample).
This ability of the SEM to generate surface images at sub-nanometer resolutions has been one of the key factors in the rapid development of nanotechnology.
In particular, it had a tremendous impact on the microchip industry, where it aided improving the production process at an exponential rate in accordance with Moore’s law.
Nowadays, microchip production lines heavily rely on SEM scanners to monitor and analyze the manufacturing process, that is, the wafer is scanned in between various processing steps, detecting defects, and therefore, increasing yield.

\begin{figure}[t]
    \centering
    \includegraphics[width=\columnwidth]{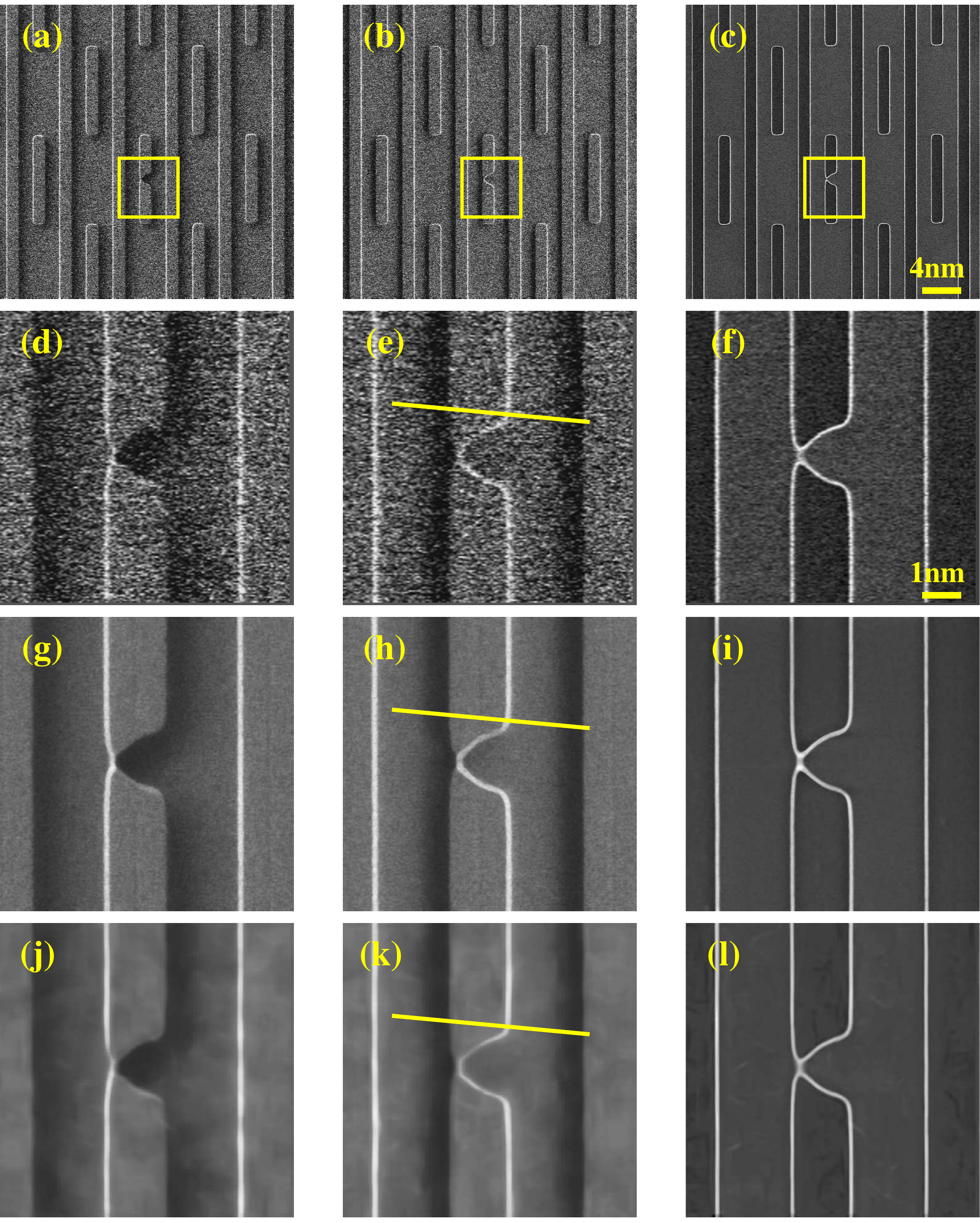}
    \caption{
        \textbf{SEM microchip images in low, high, and super-resolution.} Top row: scanned low resolution (LR) images $(20\,\mu\text{m}\times20\,\mu\text{m})$, taken from left (a), right (b), and top (c) perspectives. The yellow square shows the region of interest for enhancement with reduced field-of-view $(5\,\mu\text{m}\times5\,\mu\text{m})$. The second row (d−f) shows the corresponding regions of interest from the LR images, enlarged. The third row (g−i) shows the images acquired using the SEM high resolution (HR) mode, scanning the same areas chosen by the yellow squares above. The bottom row (j−l) depicts the enhanced super-resolution (SR) images recovered from their LR counterparts (second row). The yellow lines displayed in (e), (h), and (k) are used for the
        line cuts analyzed in \cref{fig:figure3}.
    }
    \label{fig:figure1}
\end{figure}

Two main parameters control SEM imaging: scanning velocity and resolution.
Naturally, fast scanning leads to reduced quality and added noise, while resolution determines the physical size of a pixel.
Both parameters impact acquisition time.
As microchip dimensions are reduced in accordance with Moore’s law, the complexity of the fabrication process increases.
This results in a need to scan with ever smaller pixel sizes and over a larger number of locations on the chip, in order to detect and classify defects and monitor the process.
Two common modes of SEM operation are scanning a large field of view, employing a fast acquisition time yet low resolution (LR), or slowly scanning a small field of view, achieving a higher resolution (HR) and better image quality.
An example of both scanning modes for three different perspectives can be seen in \cref{fig:figure1}.
HR scans are considered costly since their acquisition time is lengthy and their field of view is small.
Scanning the same area in HR mode takes about two orders of magnitude longer.
On the other hand, high resolution is often a necessity in the semiconductor manufacturing industry, as the ever-increasing demand for faster and more energy efficient microchips is shifting the industry toward smaller structures and more complex manufacturing processes, pushing current SEM technology to its limit.
Consequently, there is great interest in scanning large areas quickly, while maintaining high resolution.

These goals, which are clearly conflicting in current SEM technology, inspired our approach of finding a solution that breaks the immediate link between the inherent physical scanning resolution of the SEM and the resolution of the output image.
Here, we present an algorithmic method operating on a low-resolution SEM image obtained in a fast scan, to enhance its resolution and image quality up to the level of the HR image obtained in a slow scan using the same system.
Our approach consists of two steps: an offline stage of learning a joint dictionary from a sequence of LR and HR images of the same region in the chip, followed by a fast-online super-resolution (SR) step where the resolution of a new LR image is enhanced.
The SR process can be simply described as changing the mathematical representation over which the LR image is spanned from a LR basis to a suitable HR one.
Both steps are based on concepts of sparsity: the fact that SEM images have a characteristic structure, and hence they can be represented compactly in some mathematical basis that is learned in the first (dictionary learning) step.

% TODO: Update refs
The concept of sparsity lies at the heart of our approach and has been used extensively in signal processing, statistics, computer vision, and more.
According to sparse approximation theory~\cite{Eldar2015,Elad2010}, signals can often be faithfully represented as a linear combination of just a few elements from a given basis, that is, a dictionary.
This observation is true for many classes of natural images, and in principle for microchip SEM images, as we demonstrate here.
Sparse models underlie many audio and image compression (MP3, JPEG, JPEG2000) methods, as well as many state-of-the-art image processing techniques.
In general, assuming a sparse representation allows isolating the important information out of a high-dimensional signal, while removing uninformative noise.
In the past decade, the field of compressed sensing (CS)~\cite{Candes2006a,Candes2005,Donoho2006a,Eldar2012} has emerged.
CS enables the recovery of signals from partial and noisy measurements, by exploiting sparsity.
The mathematical framework provided by CS assures that a sparse signal can be exactly recovered from just a small set of linear measurements~\cite{Candes2006a}, given a known dictionary and a known measurement (sensing) system.
Relying on these principles, the concepts of sparsity and CS have been used for a variety of applications, ranging from sub-Nyquist sampling~\cite{Mishali2010,Mishali2011} with applications in radar~\cite{Bar-ilan2014,Baransky2014} and ultrasound~\cite{Tur2011,Wagner2012,Chernyakova2014}, to super-resolution imaging~\cite{Gazit2009,Szameit2010}, phase retrieval~\cite{Shechtman2011,Sidorenko2015,Shechtman2015}, ankylography~\cite{Mutzafi2015}, mapping the coherence function of light~\cite{Tian2012}, holography~\cite{Rivenson2013}, single pixel camera~\cite{Duarte2008}, ghost imaging~\cite{Katz2009}, and even quantum state tomography~\cite{Mirhosseini2014,Howland2014,Ren2016,Ren2017}.
These sparsity-based ideas inspired our current work.

The task of building a dictionary over which a family of signals, sharing the same structural features, is sparsely represented was extensively studied in recent years and is referred to in the literature as dictionary learning~\cite{Aharon2006} (DL).
The first step of any DL method is obtaining a large enough training set of measurements from the same family of signals, in our case LR and HR patches of SEM microchip images.
Choosing a complete and representative training set allows DL algorithms to construct a compact dictionary which contains only the most basic building blocks that compose the signal family.
A correctly designed dictionary can then approximate the structure of signals from within the family well.

The DL step is the most important and complex step in our methodology. One of the reasons for the success of our approach, is that we construct dictionaries that contain paired elements of both high and low resolution (the elements are called ``atoms'' in the context of DL).
This specification allows each dictionary atom to naturally map a LR patch to its HR counterpart.
In addition, we leverage different perspectives acquired simultaneously by the SEM, as will be explained below.

In our model, each dictionary atom is composed of a LR patch concatenated with a HR twin patch.
Choosing a patch size that is too small would result in dictionary atoms that cannot contain the unique inherent structure of the microchip, while specifying a large patch size increases the computational burden.
Hence, the patch size should be chosen according to the data at hand (see more details in the Supporting Information).
For our data, we use a patch size of $23\times23$ pixels.
The next design consideration is to determine the size of the dictionary, namely, the number of atoms in it.
In our specific implementation, the number of dictionary elements is defined as 3−5 times the number of coefficients in each atom (which in our case is the patch size $23^2=529$).
Thus, there is a built-in redundancy in the dictionary, since it has more elements than the intrinsic degrees of freedom. This is needed to ensure the successful description of each microchip patch as a sparse combination of the dictionary atoms.

Once the desired dimensions of the dictionary are set, the next step is to obtain and organize the training data for DL.\@
In our tests, we find that a relatively small set comprising of just several duos of scanned images of both low and high resolution (5 in our case), acquired from the exact same area in the microchip, suffices.
From these pairs of LR and HR images we can extract a multitude of patches of appropriate sizes.
The patches are simultaneously extracted and stored as part of the training set.
The extraction process is randomly performed from the available LR and HR training images, followed by a test for each of the extracted patch duos that ensures it contains a significant part of an actual microchip pattern.
The testing prevents us from including patches that contain mostly background noise into the dictionary training set.
Further details on the patch selection and pairing process are provided in the Supporting Information below.
The training set should be much larger than the number of atoms in the dictionary, roughly a hundred times bigger.
In our case the dictionary is composed of $2048$ atoms, while the training set consists of $250,000$ samples.
This ensures a thorough learning stage, which yields a robust dictionary, able to faithfully describe many possible microchip features.

The collected training data is the input to the DL algorithm.
In what follows, we use a modified version of the K-SVD~\cite{Aharon2006,Rubinstein2008} method, yet we emphasize that any DL technique could be adapted to this task.
Our algorithm uses an iterative approach to find an approximate solution to the following non-convex minimization problem~\cite{Aharon2006}:
\begin{align}
    \label{eq:DictionaryLearning}
    \left[ D , X \right] =& \arg\min_{D',X'}\norm{T-D'X'}_F^2\,, \\
        &\text{subject to } \norm{x_j}_0 < k_0, \quad \forall j \,. \notag
\end{align}
Here, $T\in\R^{n \times N_T}$ is a constant matrix containing a training sample in each of its columns, with $n$ pixels in each sample and an overall $N_T$ number of samples.
The matrix $D'\in\R^{n\times N_D}$ represents the dictionary where each of its $N_D$ columns is an atom, and $X'\in\R{N_D\times N_T}$ is a variable sparse matrix, containing the representation coefficients that span $T$ over the dictionary.
The zero-norm function $\norm{\cdot}_0$ counts the number of nonzero elements in its argument.
The constraint $\norm{x_j}_0<k_0$ enforces that every single training sample from T is described by no more than $k_0$ atoms from the dictionary (further details on the selection of $k_0$ are provided in the Supporting Information below).
Solving \cref{eq:DictionaryLearning} with the K-SVD algorithm yields two output arguments.
The first is the dictionary $D$, containing the atoms that sparsely represent the training set $T$.
Once the dictionary is obtained, the algorithm computes the sparse representation $X$, such that $T \approx DX$.

\begin{figure}[t]
    \centering
    \includegraphics[width=\columnwidth]{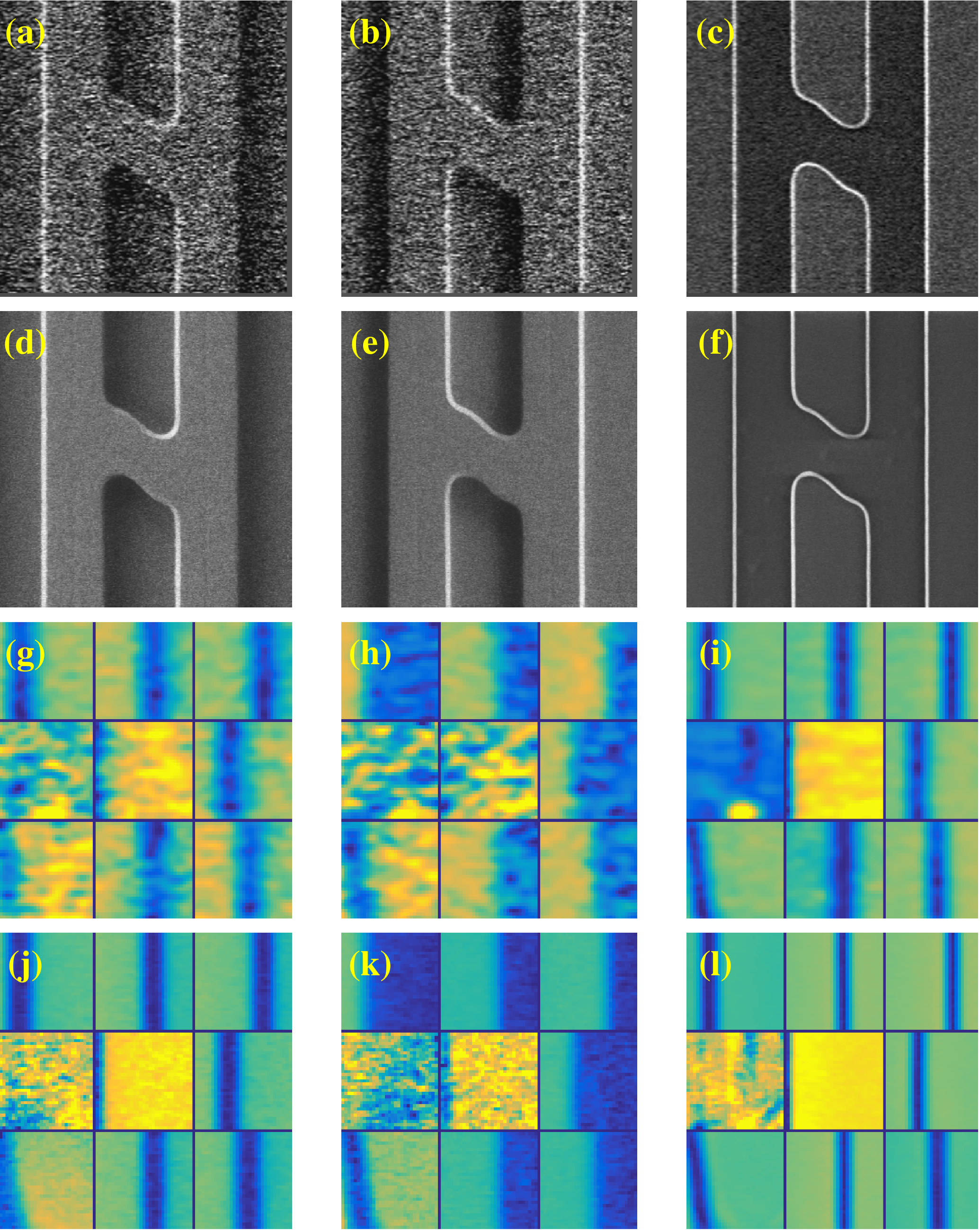}
    \caption{\textbf{Dictionary atoms mapping.}
        The top and second rows (a−f) show the LR and HR SEM images in the three perspectives, used to assemble the dictionary training set.
        The two bottom rows display subsets of nine ordered and paired atoms from each trained dictionary. The atoms in the LR trained dictionaries (g−i) are clearly mapped to the HR atoms in (j−l).
        This mapping connects every LR feature to a HR one and acts as a key enabler for the super-resolution methodology.}
    \label{fig:figure2}
\end{figure}

We adapt \cref{eq:DictionaryLearning} goals, by expanding the training set and dictionaries into concatenated matching pairs of low and high resolution, namely, solving
\begin{align}
    \label{eq:DictionaryLowHigh}
    \left[ \begin{pmatrix} D_\ell \\ D_h \end{pmatrix} , X \right] =&
    % {\arg\min}_{D'_\ell,D'_h,X'} \norm{
    \argmin_{D'_\ell,D'_h,X'} \norm{
        \begin{pmatrix} T_\ell \\ T_h \end{pmatrix} - 
        \begin{pmatrix} D'_\ell \\ D'_h \end{pmatrix}X' }_F^2\,, \\
    &\text{subject to } \norm{x'_j}_0<k_0,\quad\forall j\,. \notag
\end{align}
From here on, the subindex $h$ denotes an HR term, and the subindex $\ell$ refers to a LR term.
The training set ${T=\binom{T_h}{T_\ell}\in\R^{2n\times N_T}}$, used as input for the DL process in \cref{eq:DictionaryLowHigh}, consists of concatenated and paired LR and HR elements, thus ensuring that the trained dictionaries ${D_\ell\in\R^{n\times N_d}}$ and ${D_h\in\R^{n\times N_d}}$ are fully synchronized and paired as well.
An example for the pairing established between the atoms within each dictionary and an image duo used in the creation of the training set are shown in \cref{fig:figure2} and will be further explained below.
The accurate mapping between the dictionaries is an essential requirement that paves the way towards the next step, the actual SR reconstruction.

After a dictionary is learned, it is used to perform SR.\@
This stage consists of taking a new LR image acquired by the SEM (but was not part of the training set used to construct the dictionary) and enhancing its resolution by relying on the feature mapping established between low and high-resolution samples in the DL stage.

The SR reconstruction process requires no prior information, apart from the dictionary, constructed in the learning stage.
This process begins by taking the LR image and interpolating it to the HR grid as seen in \cref{fig:figure1}, using standard interpolation techniques (see details in the Supporting Information below).
Although the interpolated LR image resides on the same grid as the HR image, its spectral content has an inherent cutoff frequency, determined by the ratio between low and high resolutions.
Next, the interpolated LR image is divided into an ordered set of patches.
Each LR patch is stored in a matrix ${Y_\ell\in\R^{n\times N}}$, where $n$ is the number of pixels in each patch, and $N$ is the overall number of patches that compose the full LR image.
After decomposing the image, we solve the following minimization problem for obtaining the sparse representation $X^*$ over the dictionary $D_\ell$:
\begin{align}
    \label{eq:X_star}
    X^* =& \argmin_X \norm{Y_\ell-D_\ell X}_F^2\,, \\
    &\text{subject to } \norm{x_j}_0<k_0, \quad \forall j \,. \notag
\end{align}
The information encompassed in $X^*$ is now used to span the microchip at a higher resolution by using it in the overcomplete basis expansion of the HR dictionary via
\begin{align}
    \hat{Y}_h = D_h X^* \,.
    \label{eq:SR_Reconstruction}
\end{align}
This step has low computational complexity, since both the dictionary size is small and only a few of the coefficients in each column of $X^*$ are different from zero.
After calculating the matrix $\hat Y_h$, the patches contained in it are stitched together to form the final SR image.

Example sets of LR, HR, and SR images are displayed in \cref{fig:figure1}, the SR images are sharper and much cleaner, as a result of completing the missing spectral information and removal of the unwanted noise.
Enhancing the resolution of SEM images is essentially extrapolating the unknown high-frequencies missing from the LR images, while removing much of the noise introduced during the acquisition.
That is, what we demonstrate here is sparsity-based bandwidth extrapolation, based on dictionary learning.

In our experiments, we solve \cref{eq:X_star} using a fast greedy algorithm, specifically a fast variation of orthogonal matching pursuit (OMP)~\cite{Pati1993,Davis1997}, although any other sparse recovery technique can be used as well. The OMP algorithm detects, for each patch from the LR image, the most correlated dictionary atoms from within the LR dictionary $D_\ell$. It then finds the best representation for the patch using just those few selected atoms. Here, usually only 3−5 atoms are needed to faithfully describe the information contained in each patch. The OMP algorithm is fast by nature since it is a greedy algorithm: in each step OMP adds a single element to every column of $X$, which is the single best atom to sparsely describe a given patch. If the difference between the sparse combination of the already found atoms and the original patch does not meet a predetermined tolerance (dependent on the noise level of the LR image), another atom is added to minimize the difference. Consequently, the number of nonzero elements in each column of $X$ is incrementally growing, one by one in each iteration of

the OMP, until either the tolerance is achieved, or the limit of $k_0$ nonzero terms is reached.
For computational efficiency, we solve \cref{eq:X_star} by using the BatchOMP~\cite{Rubinstein2008} extension, which solves for all the columns of $X$ in parallel.
The sparse representation stored in the matrix $X^*$ now contains all of the intrinsic information on the scanned sample and allows spanning it at higher resolution, while removing much of the noise.

The formulation in \cref{eq:X_star,eq:SR_Reconstruction} considers just a single perspective, acquired by the SEM in the LR mode, to recover the SR image.
We can further improve performance by exploiting the fact that all different perspectives are acquired simultaneously from the exact same area of the scanned specimen.
For generalizing the step of DL to multiple perspectives, we would like to train dictionaries for both low and high resolution for every available perspective (three perspectives in our case), for a total of six paired dictionaries.
To produce these dictionaries, paired training samples are simultaneously extracted for all the perspectives and resolutions, generating the training sets ${T^i = \binom{T_\ell^i}{T_h}}$, for each $i$th perspective $(i = 1, 2, 3)$.
It is imperative that the training samples are extracted from the same respective areas for all of the perspectives, so that all of the sets are matched, leading to paired dictionaries.
The DL process is performed again using the K-SVD method, but now solving a bigger concatenated DL problem that takes into account all of the available information simultaneously:
\begin{align}
    \label{eq:Multi_DL}
    \left[ \begin{pmatrix*}D^1\\D^2\\D^3\end{pmatrix*} , X \right] =&
            \argmin_{D^{1'},D^{2'},D^{3'},X'}
            \norm{\begin{pmatrix}T^1\\T^2\\T^3\end{pmatrix}-\begin{pmatrix}D^{1'}\\D^{2'}\\D^{3'}\end{pmatrix} X' }_F^2\,, \\
        & \text{subject to } \norm{x'_j}_0<k_0, \quad \forall j\,. \notag
\end{align}
Here $D^i = \binom{D_h^i}{D_\ell^i}$ are the high and low-resolution dictionaries for the $i$th perspective.
\Cref{eq:Multi_DL} guarantees that the dictionaries for all perspectives and resolutions are paired, as seen in \cref{fig:figure2}.

\begin{figure}[t]
    \centering
    \includegraphics[width=\columnwidth]{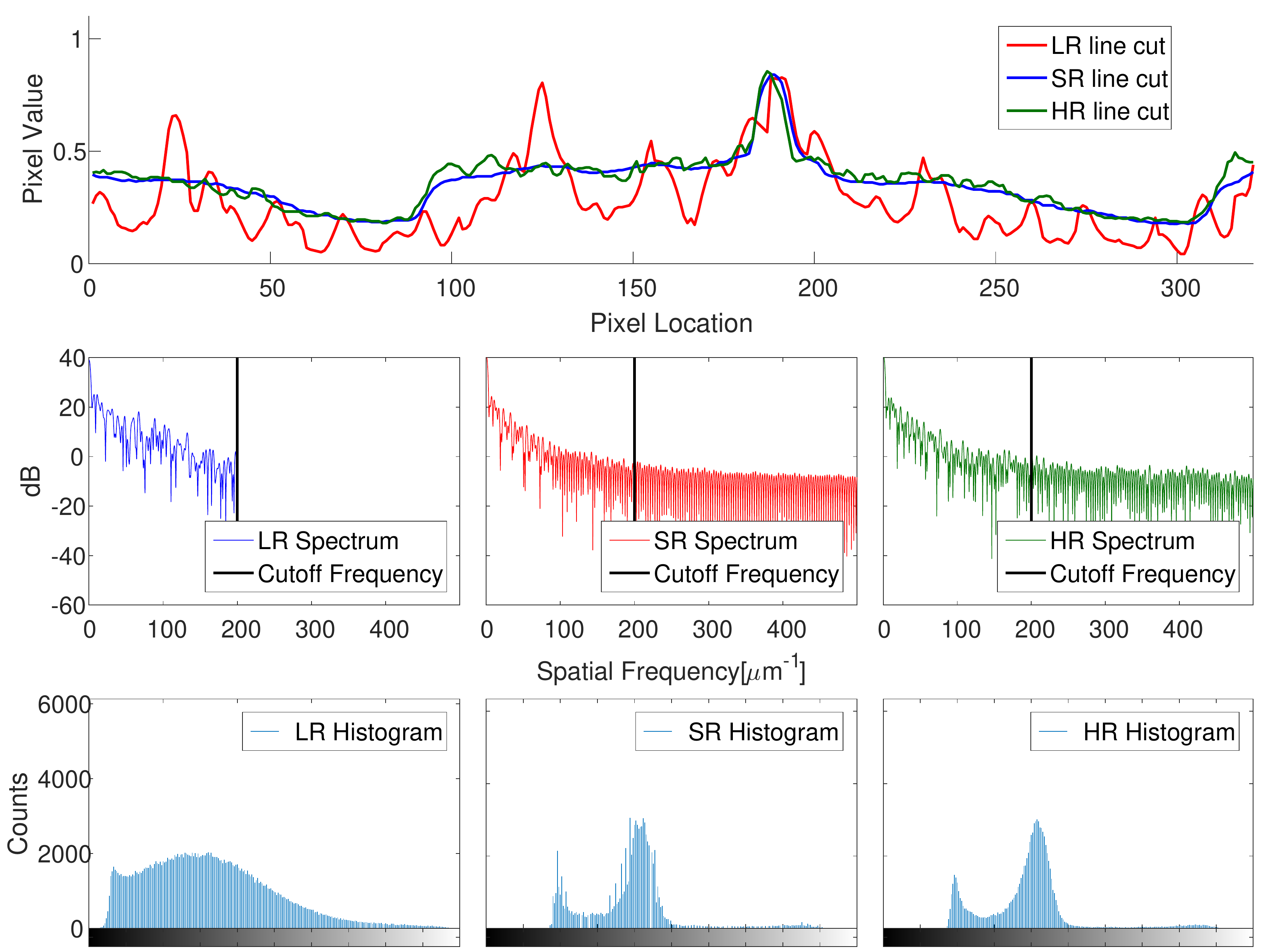}
    \caption{\textbf{Performance analysis.} The top row shows the values of the yellow line cuts from \cref{fig:figure1}, for the three images (LR, SR, and HR). The SR line cut is an almost perfect reconstruction of the HR line, except for considerable reduction of noise. The LR cut contains severe errors (e.g., the peaks at pixels $25$ and $125$) that are removed by our method, since the algorithm uses the shared information from the different perspectives simultaneously to counter gross errors inflicted by high noise. The middle row displays the 1D discrete Fourier transform of the line cuts, with the maximal spatial frequency of the LR image marked by a vertical bold line, emphasizing that 60\% of the spectral content of the HR image is extrapolated. The bottom row shows the histograms of the whole respective images from which the line cuts are taken, highlighting that the statistical nature of the SR image is well approximated. The horizontal axis depicts the gray intensity values, from black to white, of the pixels in the image, while the vertical axis displays the number of pixels in each intensity bin (divided into 512 gray level bins).}
    \label{fig:figure3}
\end{figure}

The sparse pursuit and reconstruction steps are also generalized to account for the information shared between the different perspectives.
By modifying \cref{eq:X_star} the sparse representation that spans all perspectives is obtained at once:
\begin{align}
    \label{eq:Multi_X_star}
    X^* =& \argmin_X \norm{\begin{pmatrix}Y^1\\Y^2\\Y^3\end{pmatrix}-\begin{pmatrix}D_\ell^{1'}\\D_\ell^{2'}\\D_\ell^{3'}\end{pmatrix} X' }_F^2\,, \\
    &\text{subject to } \norm{x'_j}_0<k_0, \quad \forall j\,. \notag
\end{align}
where now ${X^*\in\R^{3N_D \times 3N}}$. The reconstruction relies on the combined sparse representation $X^*$, via
\begin{align}
    \label{eq:SR_Multi}
    \begin{pmatrix}\hat Y^1_h\\\hat Y^2_h\\\hat Y^3_h\end{pmatrix}=\begin{pmatrix}D_h^{1}\\D_h^{2}\\D_h^{3}\end{pmatrix} X^*\,.
\end{align}
The three different perspectives are taken at the exact same time and location, so that the resulting measurements have the same sparse representation $X^*$ over different dictionaries, as implied by \cref{eq:Multi_X_star}.
Joint processing improves the signal-to-noise ratio by increasing the number of equations. In addition, since noise from different perspectives is uncorrelated, errors resulting from outliers in a single perspective are mitigated, as will be further discussed below, and illustrated in \cref{fig:figure3}.

Our experimental data is produced by an Applied Materials SEMVision G6 Semiconductor Defect Review (DR) tool.
The resolution of the SEMVision G6 is $1.4\,\text{nm}$, the best resolution of any DR tool available in the industry to date.
The images are of various semiconductor microchip samples whose features range between sizes of $28$ and $10\,\text{nm}$.
Respectively, the field of view of all LR images is between $20$ and $2\,\mu\text{m}$, while that of the HR images is $5$ to $0.5\,\mu\text{m}$. HR scanning takes significantly longer than LR scanning, because the scan is performed over a significantly finer grid, involving more measurements for imaging the same area.
In addition, for producing a high-quality HR scan, each grid point is scanned multiple times, and the result is averaged to increase SNR.\@
Furthermore, in the SEM setup we use, the scanning beam width and noise statistics are considered the same for LR and HR scans.

Typical results of recovering HR SEM images from their LR versions are displayed in \cref{fig:figure1}.
The top row depicts the entire field of view $20\,\mu\text{m}\times20\,\mu\text{m}$ of low-resolution scans acquired simultaneously from three perspectives of the same microchip.
The second, third, and fourth rows depict a magnified region ($\times2.5$) of a narrower field of view $5\,\mu\text{m}\times5\,\mu\text{m}$, marked by the yellow square in the top row.
The goal is to enhance the resolution of these LR images to the level of their respective HR images, while removing the noise.

We next test our methodology on real SEM data. The first stage is to construct six dictionaries, one for each perspective and resolution, using the sparsity-based methodology described by \cref{eq:Multi_DL}.
The dictionaries are constructed from a training set of just five images, acquired from the same SEM at three perspectives.
Examples for images used in the DL stage are depicted in the top rows of \cref{fig:figure2}.
Importantly, the images used for the training stage are different than those we use later for enhancing resolution, although they are acquired in the same SEM and from the same perspectives as the data whose resolution we enhance.
A small subset of ordered atoms from each dictionary is displayed in the two bottom rows of \cref{fig:figure2}, where the pairing (LR and HR) between the atoms of different dictionaries is clearly visible, emphasizing the established mapping between the elements of the three pairs of dictionaries.

After completing the training stage, we apply our method to LR images newly acquired by the SEM and enhance their resolution by first applying \cref{eq:Multi_X_star} and then \cref{eq:SR_Multi}.
\Cref{fig:figure1} shows an example of one LR image at the three perspectives (second row) and their HR counterparts (third row).
The respective SR images, obtained by our technique, are shown in the bottom row.
When comparing the enhanced SR images (bottom row) to the LR images (second row), it is clear that the overall sharpness of the contours and features is considerably improved, while the noise is greatly reduced.
Comparing the SR (bottom row) to the HR images (third row) reveals that the details and edges are accurately reconstructed for all three perspectives.

It is important to examine the performance of our technique in a quantitative manner.
To this end, we examine cross sections (line cuts) of the LR, HR, and SR images, taken along the yellow lines shown in the middle column of \cref{fig:figure1}.
The line cuts are shown in \cref{fig:figure3}.
The top panel displays the values of the pixels along the cross-section line for each image.
Clearly, the SR cross section (red) is very similar to the HR one, preserving all its features, while removing small variances that occur due to noise in the HR images.
That is, apart from accurately reconstructing the HR images, the SR line demonstrates lower noise variance even when compared to the HR images.
This fact is not surprising, since the sparsity constraint we enforce during the SR enhancement fundamentally removes noise in a highly efficient manner.
Notably, even large differences, between the LR line and HR one, are mitigated almost entirely from the SR cross section.
For example, the large peak near pixel 125 in the LR cross-section is absent in the SR recovery, and indeed that peak is absent in the HR image, exemplifying the effectiveness of our technique.
In other words, our reconstruction can remove false information that appears in the LR images, even when this information contains peaks of values twice as large as the true (HR) values.
This feature has critical impact on the SR reconstruction: without it the SR image would often contain errors.
This is a result of the fact that the recovery process considers the mutual information from all perspectives at once.
Therefore, a systematic error present in a single perspective of the LR image is removed by taking into account valid information from the other perspectives.

The spectral contents of the cross-section lines are presented in the middle row of \cref{fig:figure3}, showing that the high spatial frequencies present in the HR image are extrapolated in the SR image, well beyond the cutoff frequency present in the LR spectrum (marked by a solid vertical line).
This shows that the algorithm extrapolates 60 percent of the spectral content of the HR image, effectively expanding the spatial spectrum of the LR image by a factor of two and a half.
The bottom row of \cref{fig:figure3} shows the histograms of LR and HR images taken from the same area and compares them to the histogram of the SR image, reconstructed from the LR image.
Clearly, the histograms of the LR and HR images have very different statistical nature; nonetheless, the histogram of the SR image is very similar to the HR image.

To further assess the performance of our algorithm, we compare the SR enhanced images to the HR ones, noting that the SR images used for the fidelity test are reconstructed from new LR images.
We use a data set of 30 images for the fidelity test, containing various patterns and structures.
Unfortunately, the specific test images cannot be displayed here, due to confidentiality commitments to their respective owners, as these are electronic chips used by commercial companies.

\begin{figure}[t]
    \centering
    \includegraphics[width=\columnwidth]{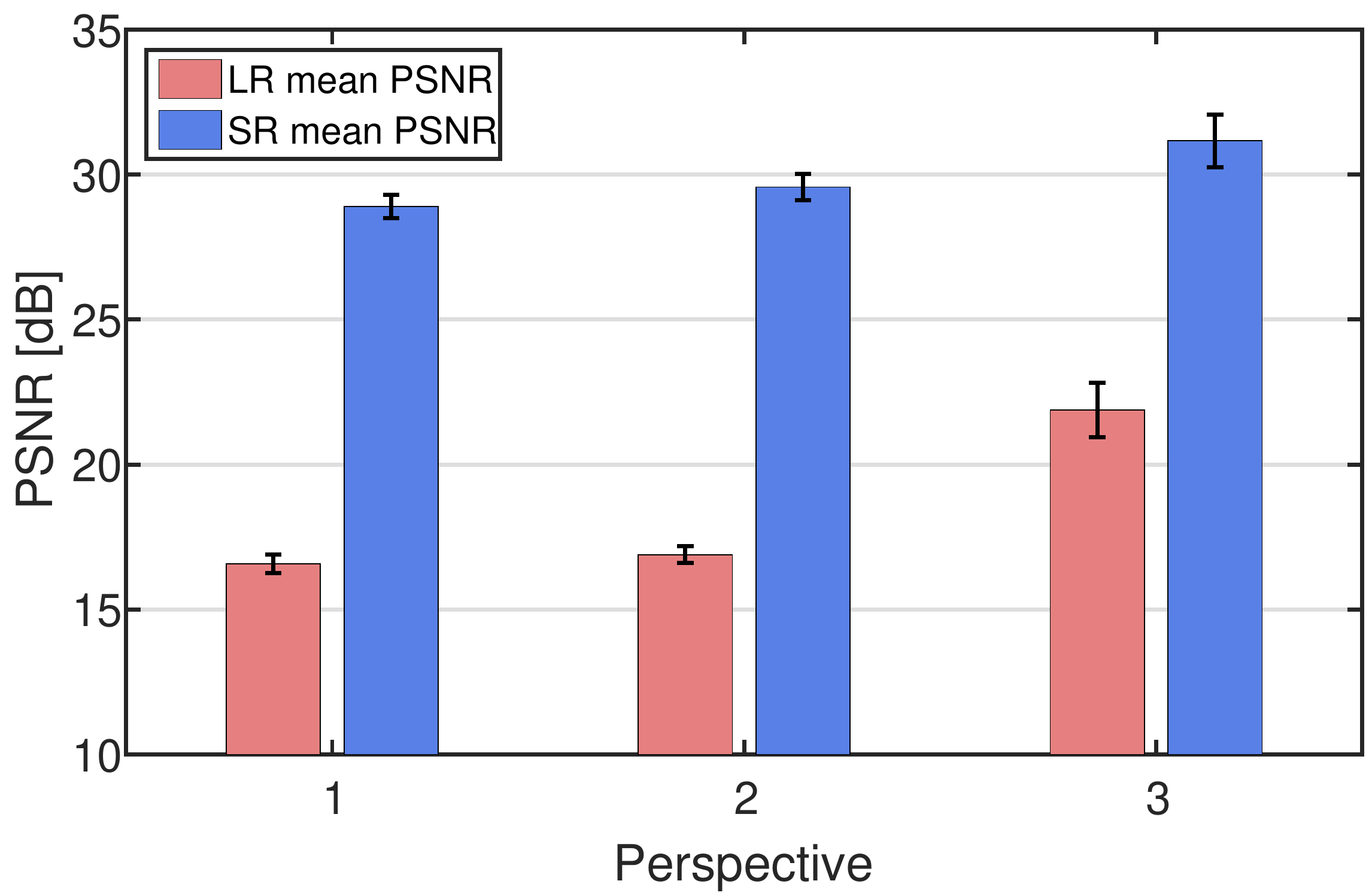}
    \caption{
        \textbf{Algorithm fidelity testing.}
        The SR algorithm is tested on 30 images that were not part of the training set used for constructing the dictionary.
        All the LR and HR data is taken from real SEM experiments (not simulated data).
        The performance is calculated using the peak signal-to-noise ratio (PSNR), using \cref{eq:psnr}, and the standard deviation of the PSNR is marked by the error bar.
        Both the LR and SR images are compared with the HR images acting as reference.
        A notable improvement of $10-15\,\text{dB}$ is achieved when comparing the SR images to their LR counterparts.
        Furthermore, the results show that performance is consistent across all perspectives throughout the image set, indicating the robustness of our methodology.
    }
    \label{fig:figure4}
\end{figure}

What we show here (\cref{fig:figure4}) is the outcome of the fidelity tests, which is meant to assess the success of our methodology.
To that end, we note that both the LR images and their HR counterparts were not included in the training set used to construct the dictionary.
Rather, the paired images are new, obtained in the same machine under conditions similar to those used to obtain the training set.
For evaluating our methodology, we consider the HR images as the ``ground truth'', although they do contain some inherent measurement noise, which naturally impedes our comparison.
The evaluation is performed using the PSNR metric, calculated as
\begin{align}
    \label{eq:psnr}
    \text{PSNR}\left( \hat Y_h,Y_h \right) = -10\log_{10} 
        \left( \frac{\norm{\hat Y_h - Y_h}_F^2}{nN} \right)
\end{align}
The performance scores of the SR enhanced set of images and their reference HR versions are displayed in \cref{fig:figure4}, as blue bars.
For comparison, the PSNR values are calculated between the LR and HR images, and their average is shown for each perspective by a red bar.
An improvement of $10-15\,\text{dB}$ is consistently gained across all images and perspectives tested.
The values achieved of approximately $30\,\text{dB}$, indicate very good reconstruction quality, especially since the reference HR images suffer from considerable measurement noise as well.

Before closing, we comment briefly on other SR approaches. 
Classical resolution enhancement methods~\cite{Irani1990,Borman1998,Park2003} relied on the prior acquisition of several LR images, each with sub-pixel shifts from one another, to generate one SR image.
Modern methods, as well as our methodology, construct an SR image from just a single LR sample.
While our solution is inspired by techniques~\cite{Yang2010,Zeyde2010,Timofte2013,Timofte2015} that combine learned dictionaries with sparse representations, nowadays, there are many alternative methods for single-image SR.\@
Among these are approaches that incorporate deep neural networks~\cite{Dong2015}, internal patch recurrence~\cite{Glasner2009,Michaeli2014}, super-resolution forests~\cite{Schulter2015,Salvador2016}, and more, that were shown to produce successful results for natural images.
Here, our work demonstrates that tailoring a modern signal processing technique to the SEM setup can improve performance to a great extent.
It is plausible that combining concepts from other modern SR methods, as those mentioned above, may further improve the results for SEM imaging.

In summary, this work is part of an ongoing attempt to incorporate cutting-edge technologies from the domain of signal processing into the realm of applied physics and optics in particular.
We proposed a simple and fast method that produces an enhanced HR image from a single noisy LR SEM scan.
Our method achieves unprecedented results in terms of reconstruction quality of SEM images while operating under high noise levels and achieving up to $\times4$ resolution enhancement in our tests (see Supporting Information for an example).
The fast-online reconstruction step accurately recovers spectral information well beyond the cutoff frequency of the image.
Our approach relies on two main foundations.
First, the exploitation of the unique structural properties of the scanned samples, specifically finding the sparse representation of the samples over a redundant mathematical basis, learned offline from available training data.
\textit{\textbf{We stress that the trained dictionaries assume no prior knowledge on the samples.}}
Second, information from several acquired perspectives is exploited to achieve good reconstruction results, even when gross errors in the measured data are present.
We show that previously lost high frequency spectral components of the LR images are extrapolated and reconstructed and that consistent results are achieved throughout our testing set, demonstrating robustness of our method.

\newpage
\section*{Supporting Information}

% Figure tags will now have an S at the start of them
\renewcommand\thefigure{S\arabic{figure}}    
\setcounter{figure}{0}

\begin{figure}[t]
    \centering
    \includegraphics[width=\columnwidth]{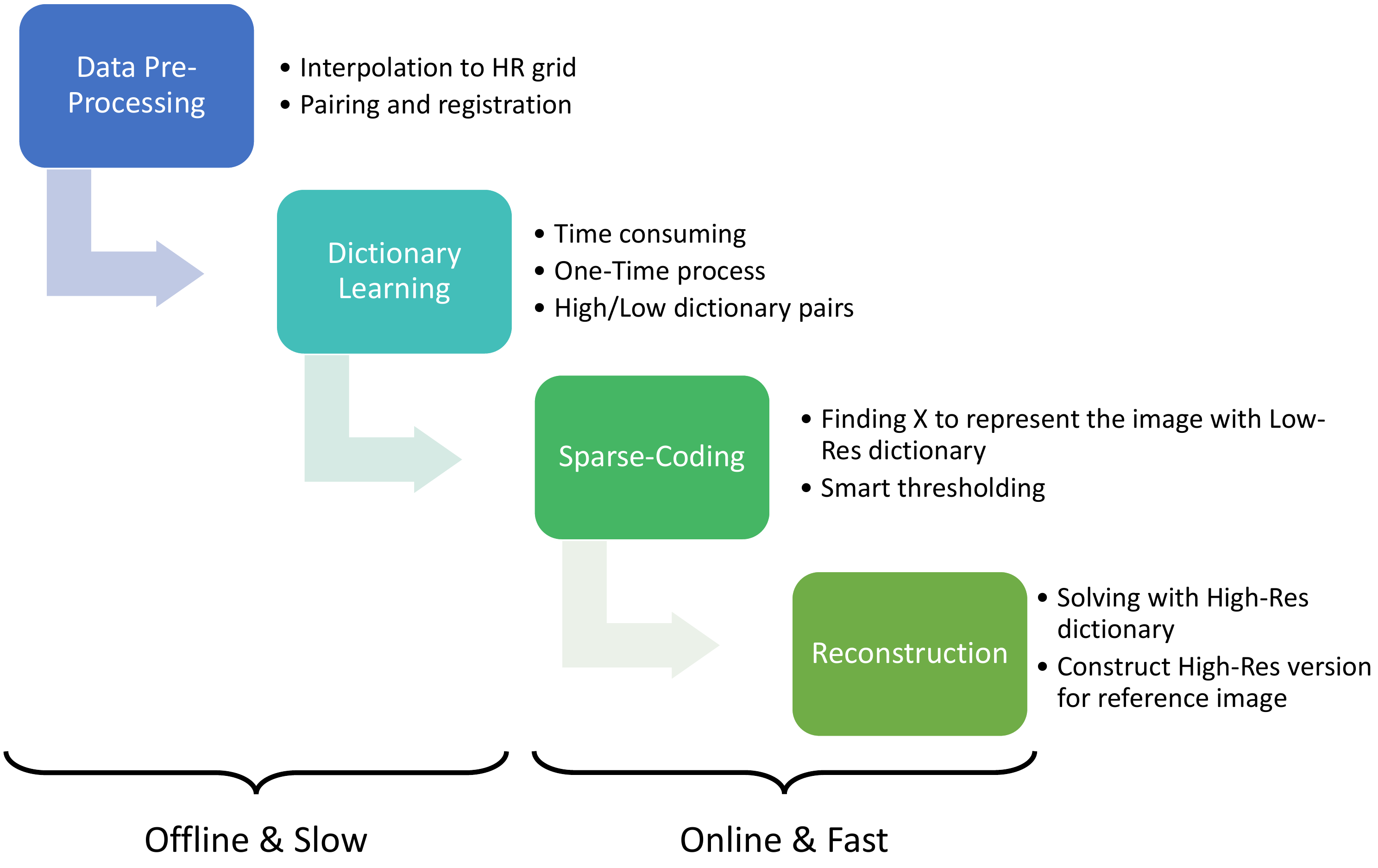}
    \caption{\textbf{Flowchart of the SR algorithm.} From left to right: the data preprocessing and DL steps are performed offline, followed by the sparse coding and reconstruction steps that are performed online to generate SR enhanced images out of single LR images. A short description, next to each block, describes the main highlights of each step, and the numbers in parentheses refer to their corresponding equations.}
    \label{fig:figureS1}
\end{figure}

In this Supporting Information, we present an overview of our super-resolution algorithm and provide additional technical details that complete the description in the main manuscript.
The dictionary learning algorithm for SEM images is detailed in Algorithm 1 below, and the online super-resolution reconstruction algorithm is described in Algorithm 2.
A flowchart diagram summarizing the main steps of the method follows in \cref{fig:figureS1}.
In addition to technical supplements, we demonstrate here the performance of our approach operating on larger zoom ratios.
In \cref{fig:fig_s3} the algorithm is demonstrated for a magnification ratio of $\times4$, and is further analyzed in \cref{fig:fig_s4}.\@
In the example presented, a total of 75\% of the spectral components of the SR image are extrapolated from the given LR image.
The SR and HR images are visually similar, and have similar statistics, as portrayed by their histograms in \cref{fig:fig_s4}.

We begin by discussing the patch size selection for the initial dictionary design.
In many common compression algorithms, such as JPEG and JPEG2000 compression, as well as in other dictionary-learning based enhancement methods \cite{Yang2010,Zeyde2010}, the patch dimensions are usually selected as $\sqrt{n}=8$ (the entire square patch is comprised of $n$ pixels).
In our experiments, we find that this size offers a good trade-off that achieves good performance in relatively noise-free images, while keeping computational complexity low.
For the SEM data at hand, which is very noisy, we find that larger patch sizes of $\sqrt{n}=19\ldots27$ pixels in each side yields better results; for the examples below and in the manuscript, we choose $\sqrt{n}=23$.
These larger patch sizes lead to better denoising, yet increase the computational complexity, which is proportional to $n$.
In the context of our SEM application, they are suited for coping with the high levels of noise apparent in the images.

Next, we consider the preprocessing performed on the HR and LR images used for dictionary learning: interpolation and registration. Specifically, the LR images (all perspectives of them) are interpolated to the HR grid by cubic interpolation, accomplished by convolution with a cubic piece-wise polynomial \cite{Keys1981}:
\begin{align}
    \tilde{y}_\ell [u,v] = \sum_{i=1}^{N-1}\sum_{j=0}^{N-1}y_\ell[i,j]w\left(\frac{u}{R}-i\right)
        w\left(\frac{v}{R}-j\right),\tag{A1}
\end{align}
where,
\begin{align}
    w(s) = \begin{cases}
        \frac32 | s |^3 - \frac52|s|^2+1,  & 0\leq|s|<1, \\
        -\frac12|s|^3+\frac52|s|^2-4|s|+2, & 1\leq|s|<2, \\
        0                                  & 2\leq|s|\,.
    \end{cases}\tag{A2}
\end{align}
Here, $y_\ell [i,j]$  is the LR image at pixel indices $\{i,j\}_0^{N-1}$ on the LR grid, and the interpolated LR image is denoted by $\tilde y_\ell [u,v] $, where ${\{u,v\}}_0^{\lfloor NR \rfloor}$ are the pixel indices on the HR grid.
The constant $R>1$ is the zoom ratio between the HR and LR image.

\begin{figure}[t]
    \centering
    \includegraphics[width=\columnwidth]{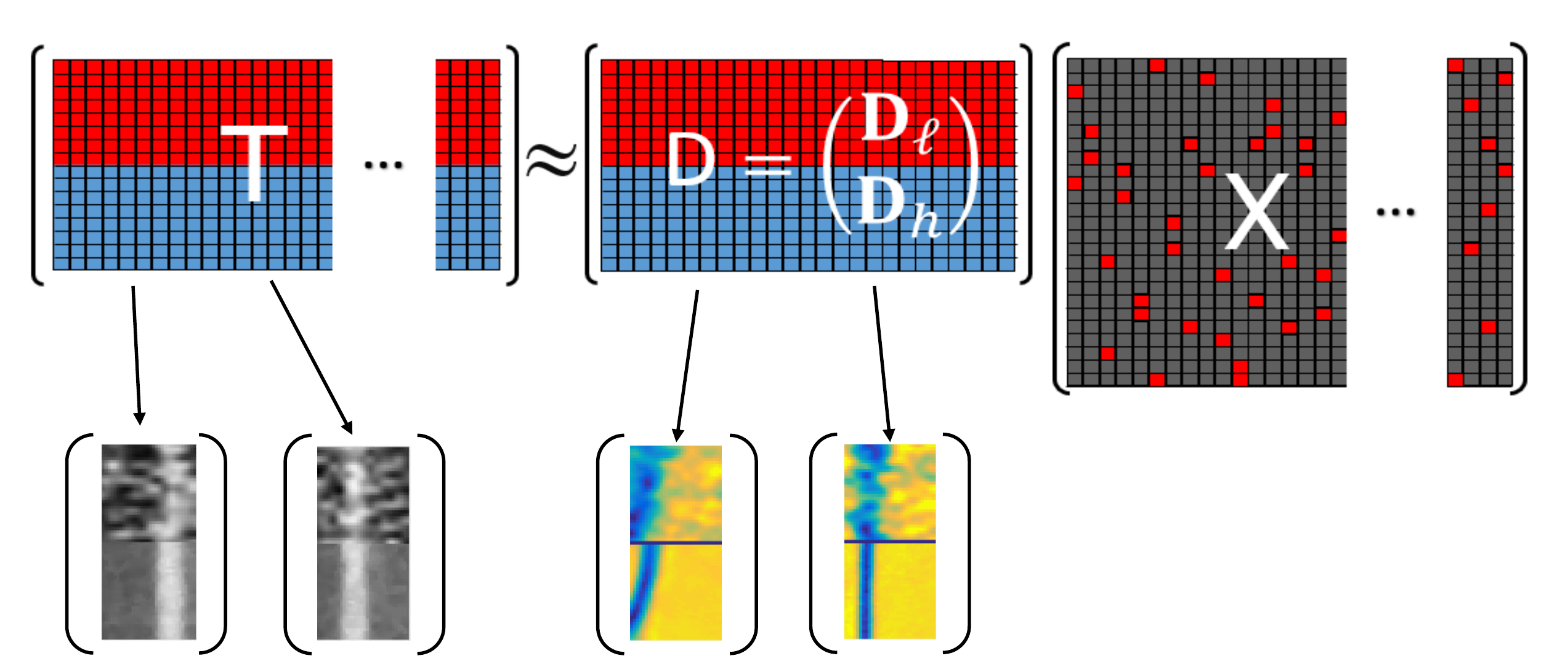}
    \caption{\textbf{Dictionary Learning Illustration for a single perspective.} The left-hand side of the equation displays the training set for a single perspective. It illustrates that each of its columns describes two paired patches, one extracted from a LR image and the other from a HR image. Correct pairing of the training set is crucial for success. On the right-hand side, each column from the dictionary describes two paired atoms, LR and HR. The atoms are paired if the training set is paired. The sparse representation $X$ multiplies the dictionary $D$, to sparsely represent the training set. Each red cell in $X$ corresponds to a non-zero coefficient while the gray cells depict zeros, emphasizing its sparse structure. For simplification purposes, the structure for a single perspective is shown, yet in practice the training samples for all perspectives are concatenated and used simultaneously.}
    \label{fig:T_DX}
\end{figure}

Registration is performed by cross-correlation, calculated between the interpolated LR image and the matching HR counterpart:
\begin{align}
    \tilde y_\ell[u,v] \gets \tilde y_\ell \left[u-\delta_1,v-\delta_2\right] \tag{A3}
\end{align}
where,
\begin{align}
    \left[\delta_1,\delta_2\right] = \arg\max_{\delta'_1,\delta'_2}\sum_u\sum_v
        \tilde y_\ell[u+\delta'_1,v+\delta'_2]y_h[u,v] \tag{A4}
    \label{eq:registration}
\end{align}
Here, $y_h$ is the reference HR image, and $y_\ell$  is the interpolated LR image that is laterally shifted until precise registration is achieved.
The lateral registration in \cref{eq:registration} is sufficient in our case, due to the fixed rotational alignment of the SEM scanner.
By first interpolating and then registering, we permit sub-pixel shifts relative to the LR grid.
This improves the precision of the registration, between the LR and HR training samples, and, consequently, leads to better paired dictionaries.
Due to the low SNR in the LR images, the cross-correlation registration may occasionally fail.
In our experiments, the cross correlation is assumed faulty if the shift values $\left[\delta_1,\delta_2\right]$, obtained in \cref{eq:registration}, satisfy ${\max⁡ \left[\delta_1,\delta_2\right] > \delta_{\max}}$, where $\delta_{\max}=6$ pixels.
In that case, we simply omit the unregistered image, since we only need a small subset of LR and HR images for training the dictionary.

\begin{figure}[t]
    \centering
    \includegraphics[width=\columnwidth]{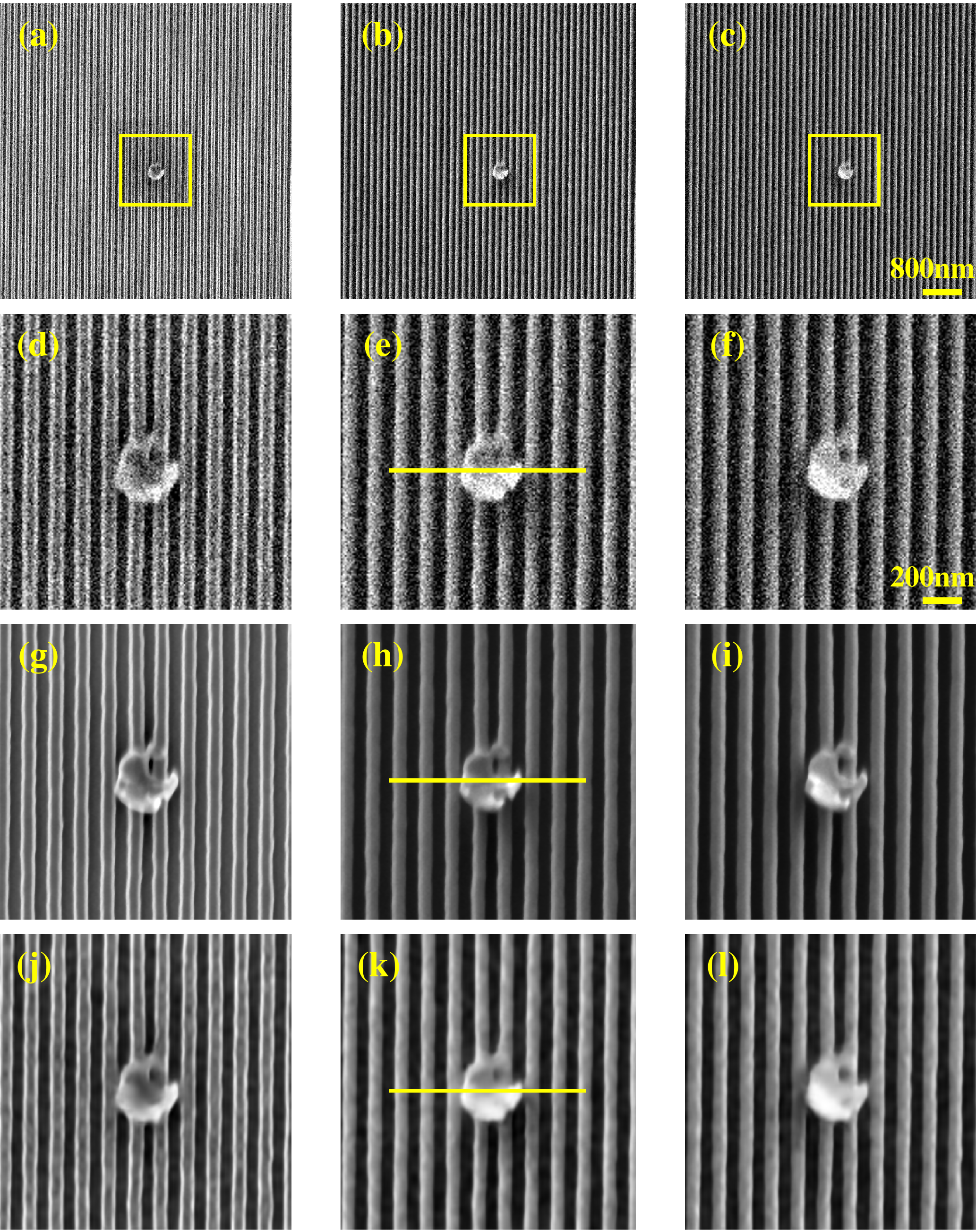}
    \caption{\textbf{SEM microchip images in low, high, and super-resolution.} Top row: scanned low resolution (LR) images $(4\,\mu\text{m}\times4\mu\text{m})$, taken from left (a), right (b), and top (c) perspectives. The yellow square shows the region of interest for enhancement with reduced field-of-view $(1\,\mu\text{m}\times1\mu\text{m})$. The second row shows the corresponding regions of interest from the LR images, enlarged. The third row show the images acquired using the SEM high resolution (HR) mode, scanning the same areas chosen by the yellow squares above. The bottom row depicts the enhanced super-resolution (SR) images recovered from their LR counterparts (second row). The yellow lines displayed in (e), (h), (k) are used for the line cuts analyzed in \cref{fig:fig_s4}.
 }
    \label{fig:fig_s3}
\end{figure}
\begin{figure}[t]
    \centering
    \includegraphics[width=\columnwidth]{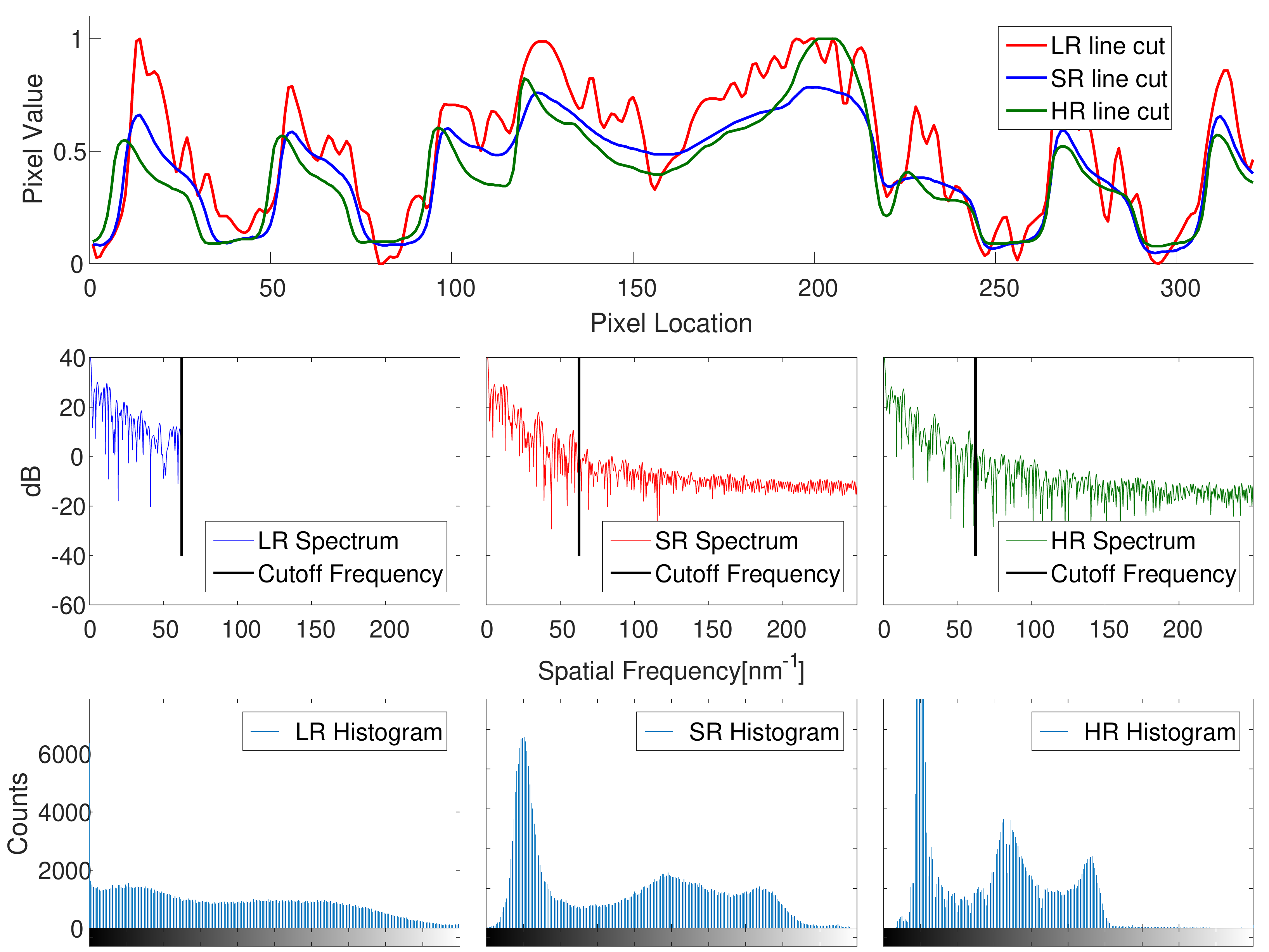}
    \caption{\textbf{Performance analysis.} The top row shows the values of the yellow line cuts from \cref{fig:figureS1}, for the three images (LR, SR and HR). The middle row displays the 1D discrete Fourier transform of the line cuts, with the maximal spatial frequency of the LR image marked by a vertical bold line, emphasizing that 75\% of the spectral content of the HR image is extrapolated. The bottom row shows the histograms of the whole respective images from which the line cuts are taken, highlighting that the statistical nature of the SR image is well approximated. The horizontal axis depicts the gray intensity values, from black to white, of the pixels in the image, while the vertical axis displays the number of pixels in each intensity bin (divided into 512 gray level bins).}
    \label{fig:fig_s4}
\end{figure}

Once the pairing between LR and HR images is complete, we extract patches for the training set.
For learning, it is beneficial to choose a set of patches that contain significant portions of microchip features, and avoid patches that contain mostly noise.
A training sample vector $t=(t_\ell^T,t_h^T )^T$ is tested by first calculating its variance $\sigma_t^2$.
If its variance is significantly higher than the variance of the background noise $\sigma^2$ (calculated by measuring the average energy in known blank areas), such that ${\sigma_t^2\geq3\sigma^2}$, then we assume that the patch $t\in\R^{2n}$ contains a meaningful portion of a microchip pattern. In that case, we add it to the training set $T\in\R^{2n\times N_T}$ as an additional vectorized column, where $N_T$ is the overall number of training samples. The training set $T$ is illustrated in \cref{fig:T_DX}, together with the trained dictionary $D=\binom{D_\ell}{D_h}\in\R^{2n\times N_D}$, and the obtained sparse representation $X\in\R^{N_D\times N_T}$.

\begin{algorithm}[t]
    \caption{Training Multi-Perspective Dual Dictionaries for SEM Images}\label{alg:train}
    \begin{algorithmic}[1]
        \Statex{\hspace{-1.5 em}\textbf{Input:} LR and HR SEM images acquired from the same perspective}
        \Statex{\hspace{-1.5 em}\textbf{Output:} Paired dictionaries for each perspective and resolution: $D_\ell^i,D_h^i\ i=(1,2,3)$}
        \State{\textbf{Interpolate} the LR images to the HR grid, and \textbf{register} the images to establish correct pairing}
        \State{\textbf{Extract} training samples from random locations from the multi-perspective images }
        \State{\textbf{Test} the training samples, and keep only viable samples }
        \State{\textbf{Solve} \cref{eq:Multi_DL} using K-SVD. For sparse-coding use BatchOMP}
        \State{\textbf{Store} the output dictionaries for the reconstruction step}
      \end{algorithmic}
\end{algorithm}

\begin{algorithm}[t]
    \caption{SR Resolution Reconstruction for SEM Images}\label{alg:superres}
    % \vspace{-1 em}
    \begin{algorithmic}[1]
        \Statex{\hspace{-1.5 em}\textbf{Input:} 
            \begin{itemize}
                \item Newly acquired LR SEM image from multiple perspectives
                \item Paired dictionaries for each perspective and resolution 
            \end{itemize} }
        \Statex{\hspace{-1.5 em}\textbf{Output:} SR reconstruction of the LR image} 
        \State{\textbf{Decompose} LR image into ordered patches for each perspective $Y_\ell^{(i)}$}
        \State{\textbf{Compute} the sparse representation $X^*$ of $Y_\ell^{(i)}$ over the LR dictionaries $D_\ell^{(i)}$,
                by solving \cref{eq:Multi_X_star} with BatchOMP }
        \State{\textbf{Generate} the SR images by $Y_h^{(i)}=D_h^{(i)}X^*$ (\cref{eq:SR_Multi})}
        \State{\textbf{Stitch} together the decomposed patches in $\hat Y_h$ to form the SR images for each perspective }
  \end{algorithmic}
\end{algorithm}

We now turn to provide more details on the sparse coding step in the online SR reconstruction, as described by \cref{eq:X_star,eq:Multi_X_star}.
This step is imperative, as it finds an approximated sparse representation $X^*$ of a sample over a given dictionary.
Before running the pursuit algorithm for a given patch out of the LR image, the patch is first tested to see whether it contains viable data.
Unless its variance $\sigma_y^2$ is larger than the background noise variance, i.e.\ $\sigma_y^2>\sigma^2$, it will be ignored and replaced by its mean value.
This test both saves significant computational time in the online step, as well as improves reconstruction results.
If a patch is found to contain a pattern, then the sparse coding step algorithm is employed to obtain its representation coefficients.
In our implementation, we stop the pursuit algorithm if the relative quadratic representation error is smaller than a pre-given constant, so that
\begin{align}
    \frac{\norm{D_\ell x^k-y}_2}{\norm{y}_2}<\epsilon, \tag{A5}
\end{align}
where $y$ is the input concatenated vector, containing paired LR patches from all perspectives, and $x^k$ is its respective sparse representation over the known dictionary $D_\ell$, at the $k$th iteration.
The relative error parameter is selected as $\epsilon=0.3$.
The upper bound for the sparse representation cardinality is selected as $k_0=\lfloor \sqrt n/2 \rfloor$, and is proportional to the square root of the intrinsic degrees of freedom in each patch ($n$).
Usually, a much smaller number of atoms is required for spanning a given patch, approximately $3-5$ atoms in our tests.
The upper bound is attained only for several patches in the entire representation.

Finally, we address the computational complexity of our online reconstruction.
The online stage is composed of two steps, sparse-pursuit and SR reconstruction.
The first is given by \cref{eq:X_star,eq:Multi_X_star}, and consists of running the BatchOMP~\cite{Rubinstein2008} algorithm for all patches that comprise the interpolated image $y_\ell$.
Its asymptotic computational complexity is given by $\O(n^2 N_D N_Y)$, where $n^2$ is the number of pixels in each patch, $N_D$ is the number of atoms in the dictionary and $N_Y$ represents the number of patches in the image.
The following reconstruction step, in \cref{eq:SR_Reconstruction,eq:SR_Multi}, has complexity $\O(k_0 N_Y )$, since it involves summing $k_0$ atoms for representing each of the $N_Y$ patches that compose the entire image.
Therefore, the overall SR reconstruction computational complexity is on the order of $\O(n N_D N_Y)$.

\footnotesize
\bibliographystyle{misc/myIEEEtran}
\bibliography{library}

\end{document}